\pdfoutput=1

\documentclass[11pt]{article}

\usepackage[preprint]{acl}

\usepackage{times}
\usepackage{latexsym}

\usepackage[T1]{fontenc}

\usepackage[utf8]{inputenc}

\usepackage{microtype}

\usepackage{inconsolata}

\usepackage{amsmath}
\usepackage{booktabs}
\usepackage{pdfpages}
\usepackage{graphicx}
\usepackage{caption}
\usepackage{multirow}
\usepackage{cleveref}
\usepackage{xspace}
\usepackage{diagbox}
\usepackage{hyperref}
\usepackage{tabularx}
\usepackage{hypcap}

\newcommand{\ex}[1]{\textit{#1}\xspace}

\newcommand{\model}[1]{\text{#1}\xspace}
\newcommand{\four}{\model{GPT-4}}
\newcommand{\threefive}{\model{GPT-3.5}}
\newcommand{\llamatwo}{\model{LLaMA2}}
\newcommand{\llamathreeone}{\model{LLaMA3.1}}
\newcommand{\gemma}{\model{Gemma}}
\newcommand{\mistral}{\model{Mistral}}

\newcommand{\debertavthree}{\model{DeBERTaV3}}
\newcommand{\true}{\model{TRUE}}

\definecolor{CadmiumOrange}{rgb}{0.93,0.53, 0.18}

\title{Evaluating Evidence Attribution in Generated Fact Checking Explanations}

\author{
 \textbf{Rui Xing\textsuperscript{1}} \ 
 \textbf{Timothy Baldwin\textsuperscript{1,2}} \
 \textbf{Jey Han Lau\textsuperscript{1}} \\  
 \textsuperscript{1}The University of Melbourne,
 \textsuperscript{2}MBZUAI\\
\texttt{ruixing@student.unimelb.edu.au, tb@ldwin.net, jeyhan.lau@gmail.com} \\
}

\begin{document}
\maketitle
\begin{abstract}
Automated fact-checking systems often struggle with trustworthiness, as their generated explanations can include hallucinations. In this work, we explore evidence attribution for fact-checking explanation generation. We introduce a novel evaluation protocol -- citation masking and recovery -- to assess attribution quality in generated explanations. We implement our protocol using both human annotators and automatic annotators, and find that LLM annotation correlates with human annotation, suggesting that attribution assessment can be automated. Finally, our experiments reveal that: (1) the best-performing LLMs still generate explanations with inaccurate attributions; and (2) human-curated evidence is essential for generating better explanations. Code and data are available here: \url{https://github.com/ruixing76/Transparent-FCExp}.

\end{abstract}

\section{Introduction}
\begin{table}[!t]
    \centering
    \small
    \begin{tabular}{p{0.95 \columnwidth}}
    \toprule
    \textbf{Claim:} In 2018, for every "share" that a photograph of a sick boy gets, Facebook will donate one dollar. \\
    ... \\
    \textbf{Evidence passage 9:} Facebook does not make donations based on the number of shares or comments a particular post or photo may receive. \\
    ... \\
    \textbf{Evidence passage 10 :} The post is an example of a long-running scam that is employed to generate user interaction by falsely claiming that shares or comments will help pay for a sick child's care. \\
    ... \\
    \textbf{Evidence passage 11 :} In 2015, the Better Business Bureau released a statement about such scams... \\
    \midrule
    \textbf{Veracity:} False \\
    \midrule
    \textbf{Selected evidence:} \{9, 10, 11\} \\
    \midrule
    \textbf{Generated Explanation:} The claim that Facebook will donate a dollar for every "share" that a photograph of a sick boy receives is false. It is crucial to note ... a particular post or photo may receive [9]. Such posts are typically part of a long-running scam ... will financially contribute to the care of a sick child [10]. Though it might initially appear benign,..., which in turn empowers them to disseminate further deceptions [11]. \\
    \bottomrule
    \end{tabular}
    \caption{An example in PolitiHop with claim $c$, veracity $v$, evidence passages $E$. $X$ is the generated explanation that cites $E$.}
    \label{tab:politihop}
\end{table}

The dissemination of online misinformation poses a significant threat to society, with consequences ranging from seeding skepticism and discrediting science, to endangering public health and safety. To verify online claims at scale, automated fact-checking systems have been proposed to classify claims based on their truthfulness~\citep{guo-etal-2022-survey}. However, debunking by simply calling a claim ``false'' can lead to a backfire effect, whereby the belief in a false claim is further reinforced rather than hindered~\citep{stephan-misinformation-2012}.

\begin{figure*}[tb]
\centering
\includegraphics[width=1\textwidth,scale=1.5]{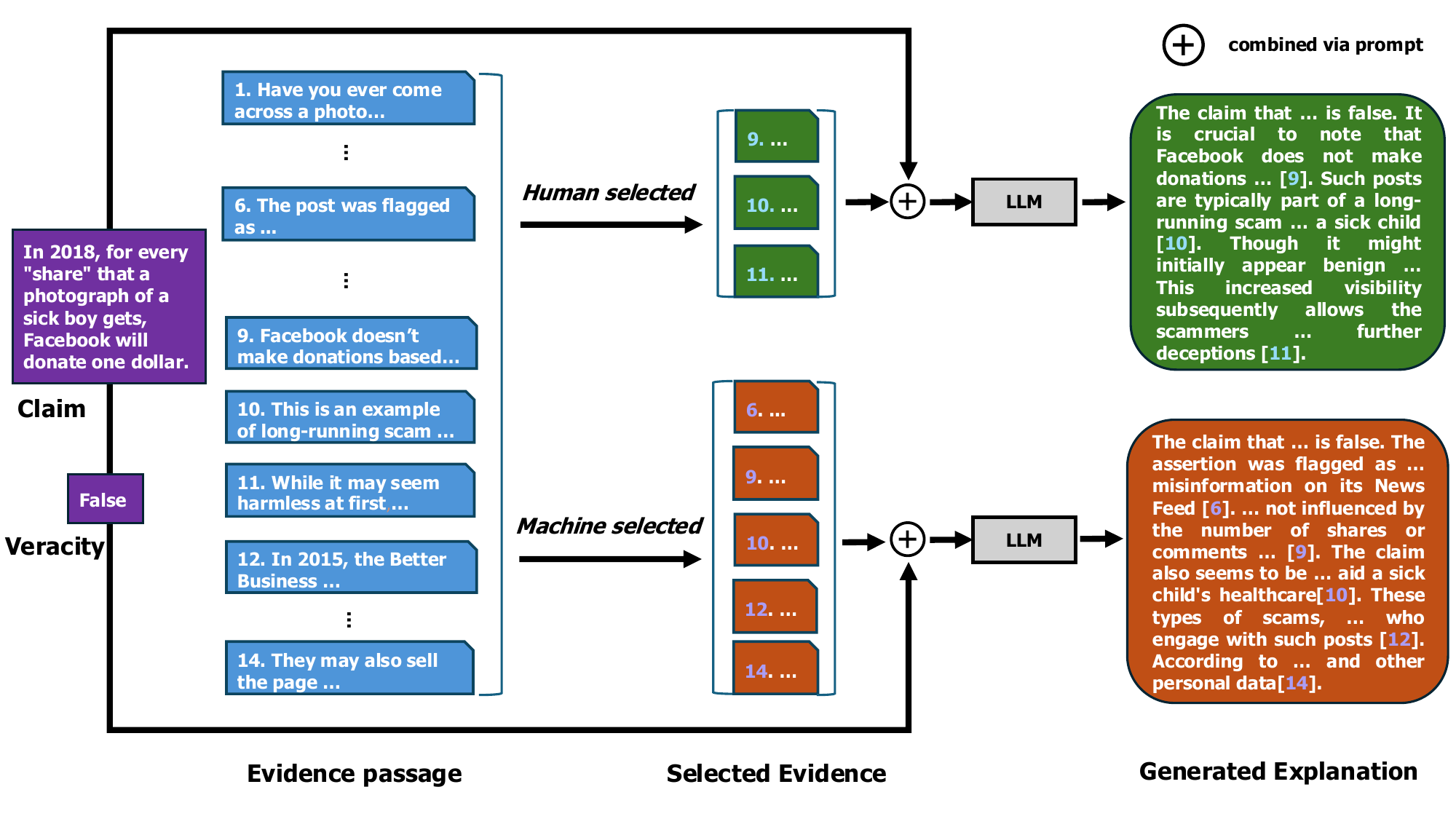}
\caption{Automated explanation generation in fact-checking. Given a claim, its veracity and a list of evidence passages, a subset of these passages is selected, either by humans or machines, and input into a large language model (LLM) along with the claim to generate the explanation.}
\label{fig:exp_gen}
\end{figure*}

As such, there's growing research on generating textual explanations to justify the outcomes of fact-checking systems using large language models (LLMs) in various contexts~\citep{eldifrawi-etal-2024-automated, wiegreffe-etal-2022-reframing}.  For automated fact-checking, the typical input is a claim and a list of retrieved evidence passages, from which a subset of evidence passages is selected and fed to the LLM for explanation generation. Table~\ref{tab:politihop} presents a typical example of the fact-checking task with a generated explanation, and \Cref{fig:exp_gen} illustrates the process of generating these explanations. 

A challenge in this task is that the explanations must provide accurate attribution of their key statements~\citep{afc_assist}. In journalistic fact-checking practice, journalists curate their sources and ensure the accuracy of every cited statement~\citep{journalism-fc-guide2022}. Machine-generated explanations should this follow convention to be trustworthy. But even if generated explanations contain attribution, how do we assess whether the attribution is faithful to the source evidence? 
Current evaluation methods are primarily based on the Attribution to Identified Source (AIS) framework~\citep{rashkin-etal-2023-measuring}, whereby an important feature is ``explicature'': the meaning of statement $s$ in linguistic context $c$. Generating explicatures typically involves either costly human annotation or a non-trivial process of decontextualization, meaning it is rarely used. Consequently, most automated evaluation methods directly use Natural Language Inference (NLI) between $c$ and $s$ to quantify AIS~\citep{gao-etal-2023-enabling, liu-etal-2023-evaluating, li-etal-2024-attributionbench}.

In this work, we ground the notion of explicature by incorporating contextual information into the evaluation process. We introduce a novel evaluation protocol -- \textit{Citation Masking} and \textit{Citation Recovery} -- to assess evidence attribution in generated explanations. The key idea is to mask the evidence citation in the generated explanation and ask annotators to recover them. The recovery process requires annotators to incorporate the linguistic context of a candidate statement during evaluation. This results in zero or more candidate statements, reflecting the complex nature of real-world attribution, and we hence frame it as a multi-label task. We first ask annotators to evaluate the attribution of a statement within the context of the full explanation. We then automate our protocol using NLI methods and LLMs as annotators. Additionally, we explore the influence of evidence selection on attribution quality, comparing human-curated and machine-selected evidence.

To summarize, our contributions are: (1) we introduce a new evaluation protocol for evidence attribution in generated explanations via citation masking and recovery; (2) we automate our protocol via LLMs and compare it with NLI models, showing that LLMs align better with human annotators; and (3) in terms of evidence source, we explore the influence of human- and machine-selected evidence on explanation generation and find that human-selected evidence produces explanations that are more accurate in terms of evidence attribution.

\begin{figure*}[t]
\centering
\includegraphics[width=1\textwidth,scale=1.5]{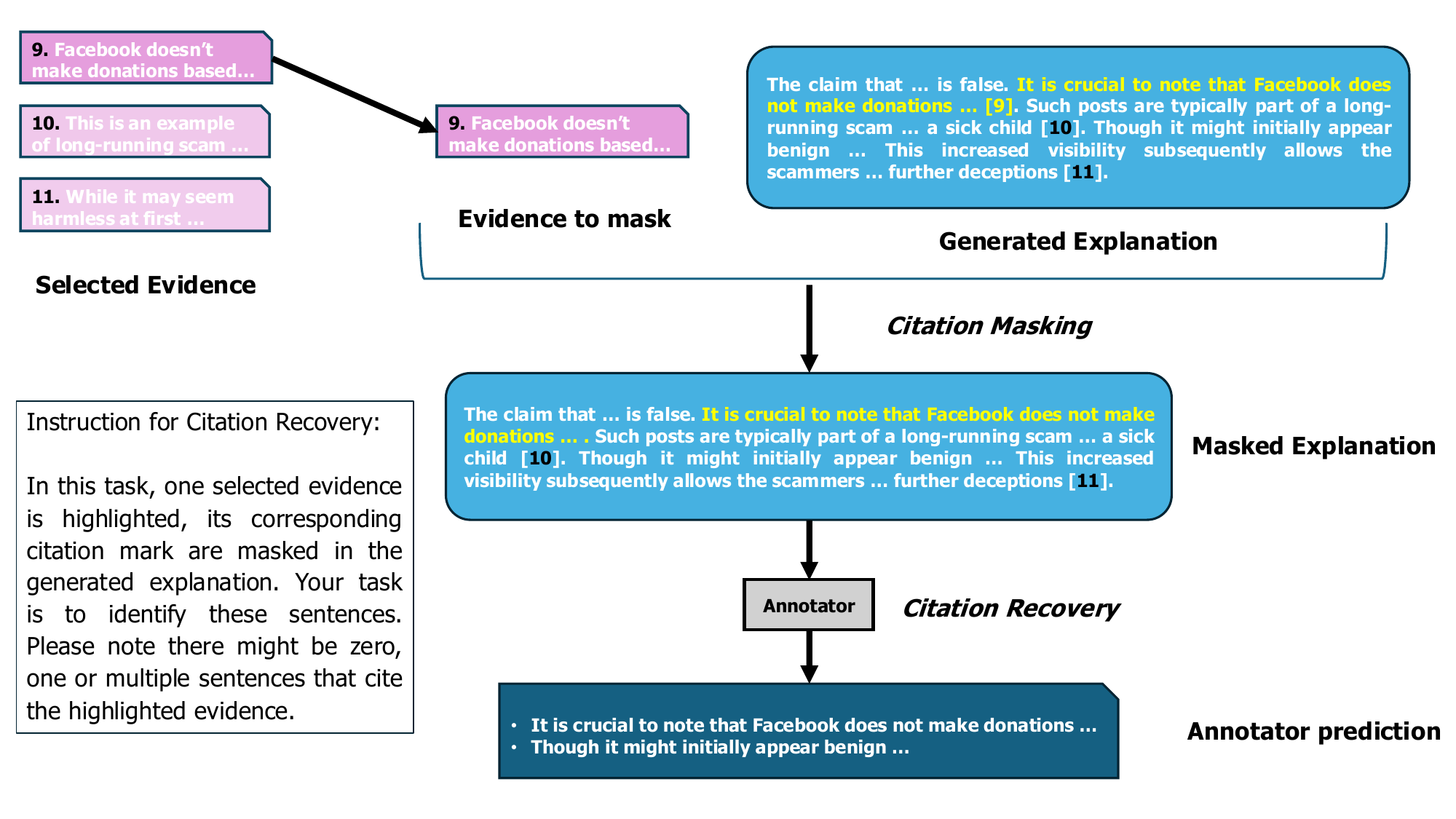}
\caption{Our human assessment protocol: citation masking and citation recovery. Given the generated explanation and a list of evidence passages, we randomly masked one sentence $e_k$ and mask its inline citation marker. Annotators are then required to perform citation recovery and predict the masked citation sentence.}
\label{fig:cit_recovery}
\end{figure*}

\section{Related Work} 
Automated fact-checking aims to classify the veracity of a claim~\citep{guo-etal-2022-survey, russo-etal-2023-benchmarking}. However, debunking solely by providing a classification label (e.g.\ \textit{False}) is neither effective nor convincing, and can induce a ``backfire'' effect where the erroneous belief is reinforced~\citep{stephan-misinformation-2012,guo-etal-2022-survey}. This motivates the task of justification or explanation generation for fact-checking. Explanation generation in the literature is framed as either extractive or abstractive summarization over evidence~\citep{atanasova-etal-2020-generating-fact,kotonya-toni-2020-explainable-automated, xing-etal-2022-automatic}. However, extractive methods struggle to produce explanations with sufficient context, and abstractive methods are prone to hallucination~\citep{russo-etal-2023-benchmarking}. 

To alleviate hallucination, most existing work incorporates citations in the generation and evaluates their attribution relative to sources~\citep{liu-etal-2023-evaluating, gao-etal-2023-enabling,  yue-etal-2023-automatic, huang-chang-2024-citation, li-etal-2024-attributionbench}. \citet{liu-etal-2023-evaluating} introduced evaluation of citation quality, and used human judges to audit the verifiability of popular LLMs. ~\citet{gao-etal-2023-enabling} developed automatic metrics along three dimensions -- fluency, correctness, and citation quality -- to assess LLMs for question answering. \citet{yue-etal-2023-automatic} explored prompting LLMs and fine-tuning smaller LMs for automatic attribution evaluation. Most existing work is based on the Attribution to Identified Source (AIS) framework~\citep{rashkin-etal-2023-measuring}, where an annotator is asked whether the statement $s$ in the explanation is fully supported by cited evidence $c$ (\ex{According to evidence $c$, $s$.}). An important assumption in AIS is explicature, an expanded statement that is interpretable out of context. Generating explicatures is a non-trivial process, often requiring costly human annotation to decontextualize statements so they can be interpreted in isolation. Most work directly adopts NLI without considering explicature~\citep{liu-etal-2023-evaluating, gao-etal-2023-enabling}. In this work, we attempt to bridge the gap by introducing a new evaluation protocol to evaluate evidence attribution in fact checking explanations.

\section{Human Evaluation of Evidence Attribution}
\label{sec:human-eval}
\subsection{Dataset}
We use PolitiHop~\citep{Ostrowski-2021-politihop} as our dataset for all experiments. \Cref{tab:politihop} shows an example instance, comprised of a \textbf{claim}, a \textbf{veracity} label, a list of \textbf{evidence passages} that provide additional relevant information about the claim, and \textbf{selected evidence} -- a subset of passages selected by \textit{human annotators} that is most relevant to the claim and collectively clarifies the veracity of the claim.\footnote{In PolitiHop, the evidence in each claim can sometimes have multiple subsets (e.g.\ \{8,10\} and \{6, 7, 8\}); for these cases we randomly select one subset for our experiments.} We sampled 100 instances from PolitiHop.

\subsection{Automatic Explanation Generation}
PolitiHop does not provide an explanation for each claim, and as such we need to generate the explanations for our experiments. We developed an LLM-based explanation generation system which assumes the following as input: (1) a claim; (2) a list of evidence passages; and (3) the veracity of the claim (e.g.\ \textit{False}). The rationale for assuming the veracity label as input is we see our explanation generation system being applied to a fact-checking classification system to provide a layer of interpretability in a real-world application. Because the set of evidence passages is typically large and not all of them are relevant, we propose a pipeline approach where we first do \textbf{evidence selection} and then feed the selected evidence, claim, and veracity label to an LLM to generate the explanation. \Cref{fig:exp_gen} illustrates the overall process.

\paragraph{Evidence Selection} 
We experiment with two methods for evidence selection: (1) \textit{Human}: we use the original human-selected evidence in Politihop; and (2) \textit{Machine}: we (one-shot) prompt an LLM to select the most relevant evidence passages (see C\ref{tab:prompt_evi_sel} for more details). Note that the evidence passages are numbered, which is important as the explanation cites them for key statements. We explore these two methods to understand the impact of this selection process -- whether hand-curated evidence is necessary for explanation generation, or if machine-selected evidence is sufficient.

\paragraph{Explanation Generation} Given a claim, its veracity label, and selected evidence (machine- or human-selected), we use a zero-shot prompt with LLMs to generate an explanation to clarify the claim. The prompt explicitly asks it to use in-line citations, making it clear that \textit{all} given evidence passages should be used (see ``Generated Explanation'' in Figure~\ref{fig:exp_gen}; the full prompt is given in Appendix Table~\ref{tab:prompt_exp_gen}). We test a range of LLMs in our experiments, noting that we always use the \textit{same} LLM for evidence selection and explanation generation. We generate explanations for all 100 sampled claims from PolitiHop.

\begin{table}[!t]
    \centering
    \resizebox{\columnwidth}{!}{
   \begin{tabular}{ccccc}
   \toprule
    \textbf{Evi Src} & \textbf{Gen} & \textbf{\shortstack{Claim  Len}} & \textbf{\shortstack{Evi  Size}} & \textbf{\shortstack{Exp  Len}} \\
   \midrule
   \multirow{3}{*}{\textit{Human}} 
   & GPT-4 & \multirow{3}{*}{22.32} & \multirow{3}{*}{3.16} & \multirow{3}{*}{140.58}\\
   & GPT-3.5 & & &\\ 
   & LLaMA2 & & &\\
   \midrule
   \multirow{3}{3em}{\textit{Machine}} 
   & GPT-4 & \multirow{3}{*}{22.32} & {5.07} & {171.26}  \\
   & GPT-3.5 &  & 4.68 & 175.89\\ 
   & LLaMA2 &  & 6.24 & 214.02 \\
   \bottomrule
   \end{tabular}
   }
   \caption{Statistics of the data and generated explanations. The ``Evi Src'' column indicates whether evidence is \textit{Human}- or \textit{Machine}-selected. ``Gen'' means the generator model. ``Claim Len'' and ``Exp (Explanation) Len'' refer to token length, which is tokenized by OpenAI's \href{https://github.com/openai/tiktoken}{tiktoken v0.7.0}. Evi (Evidence) Size refers to the number of selected evidence passages.}
    \label{tab:gen_stats}
\end{table}


\paragraph{Models}
For the evidence selector and explanation generator, we experiment with three language models: \four~\citep{gpt4}, \threefive, and \llamatwo~\citep{llama2}.\footnote{Detailed model version: \four = gpt-4-0613, \threefive = gpt-3.5-turbo, and \llamatwo = Llama-2-7b-chat. We also tested other models including LLaMA2-7B, FlanT5-xxl~\citep{flant5}, Falcon-30B~\citep{almazrouei2023falcon} and MPT~\citep{MosaicML2023Introducing} but excluded them because these models generated repetitive content with fabricated citations.} Note that we always use the same LLM for both evidence selection and explanation generation, and as such the machine-selected evidence can also be interpreted as self-selected evidence.  \Cref{tab:gen_stats} presents a statistical breakdown of the explanation generation process. In general, we can see that the machine-selected evidence set tends to be larger (i.e.\ it includes more retrieved passages), and the explanations generated using machine-selected evidence are also longer.

\subsection{Human Assessment Protocol}\label{sec:transparency} 
In this section, we introduce citation masking and recovery, a new approach for assessing the quality of evidence attribution.

\paragraph{Citation Masking}
Assume we have a claim $c$, a veracity label $v$, $m$ evidence passages $E=\{e_1,e_2,e_3,...,e_m\}$, and a generated explanation with $n$ sentences $X=\{x_1,x_2,x_3,...,x_n\}$.\footnote{Sentences are segmented with \href{https://spacy.io/}{spaCy v3.7.2}.} A subset of sentences $X_{cit}=\{x_i,x_{i+1},...,x_j\} \subseteq {X}$ contains inline citations (e.g.\ \ex{[6]}),\footnote{Technically there is a possibility that $X = \emptyset $ when the LLM fails to cite any evidence passages, but we did not see this in practice.}
where ${1}\leq{i}\leq{j}\leq{n}$. We randomly select $e_k \in E$ (${1}\leq{k}\leq{m}$) and mask its inline citation marker in the explanation (e.g.\ \ex{you are wrong [6]} $\rightarrow$ \ex{you are wrong}), producing $X^{mask}$, the masked explanation. We then denote $X^{mask}_{cit} \subseteq {X_{cit}}$ as the subset of explanation sentences where their inline citation markers are removed.

\paragraph{Citation Recovery}
We next ask annotators to recover the masked sentences. That is, annotators are presented with claim $c$, veracity $v$, full evidence set $E$, evidence passage $e_k$, and masked explanation ${X^{mask}}$, and they are asked to find all sentences that should cite $e_k$. In other words, the task is to recover $X_{cit}^{mask}$. Denoting their prediction as $X_{pred}^{mask}$, a perfect identification means  $X_{pred}^{mask} = X_{cit}^{mask}$. Note that $X_{cit}^{mask}$ sometimes contains multiple sentences (e.g.\ when the explanation has 2 sentences that cite $e_k$) so this is a \textit{multi-label} classification problem. To measure the degree of overlap between $X_{pred}^{mask}$ and $X_{cit}^{mask}$, we compute set precision, recall, and F1. A high F1 performance indicates the explanation is accurate in its attribution and faithful to the source evidence.

\paragraph{Attribution Score Computation}\label{para:attr_score}
Given an annotator prediction $X_{pred}^{mask}$ and the reference label set $X_{cit}^{mask}$, $Precision$ reflects the proportion of reference label in annotator prediction:
\begin{equation}\label{eqn:precision}
    Precision= \frac{ |X_{pred}^{mask} \cap X_{cit}^{mask} |} { |X_{pred}^{mask}| }
\end{equation}

Similarly, $Recall$ reveals the proportion of reference label recovered by annotators:
\begin{equation}\label{eqn:recall}
    Recall= \frac{ |X_{pred}^{mask} \cap X_{cit}^{mask} |} { |X_{cit}^{mask}| }
\end{equation}

$F1$ combines $Precision$ and $Recall$ via the harmonic mean.
\begin{equation}\label{eqn:f1}
    F1= 2 \cdot \frac{Precision\cdot Recall} {Precision + Recall}
\end{equation}

\subsection{Human Annotation Details}\label{sec:anno_detail}

\paragraph{Annotation Procedure}
We provide the annotator with the following information: a claim, veracity label, full evidence set, a selected evidence passage, and an explanation, and ask them to select (= highlight) sentences in the explanation that should cite the selected evidence passage (as in Figure \ref{fig:cit_recovery}). See \Cref{app:interface} for the full annotation guidelines and interface.\footnote{We also ask the annotators to rate the utility of the explanation after the citation recovery task, but we did not use these ratings for the experiments in this paper.} Note that for each claim, we only select \textit{one} random evidence passage to do citation recovery. For example in Figure \ref{fig:cit_recovery}, we select citation 9 to mask in the explanation. We address this limitation when we introduce LLM as annotator in \Cref{para:llm_annotator}, where \textit{every} evidence passage is evaluated.

\paragraph{Annotator Recruitment} 
The annotation task was done on Amazon Mechanical Turk (AMT).\footnote{\href{https://www.mturk.com/}{https://www.mturk.com/}} We applied pre-screening pilot studies to find qualified annotators. We conducted individual reviews of submitted annotations, and offered feedback to annotators to address any misconceptions or confusion about the task. Annotators who performed well in the pilot study were invited to do a final round of annotation that produced the annotated data. In order to maintain high quality throughout the annotation, we used quality control questions to identify and remove poor-performing annotators. In total, we recruited 68 annotators for the final round of annotation.

\paragraph{Quality Control} 
Quality control was implemented for both the pilot study and final study. In the pilot study, each Human Intelligence Task (HIT) contained 3/6 control questions, 2 of which were positive questions containing exactly one answer each, and 1 of which is a negative question containing no correct answer (the original answer sentence has been removed). Annotators are expected to choose ``There isn't any sentence that can correctly cite the highlighted core evidence.'' in such cases. All control questions were manually validated by the first author of the paper. An annotator who fails on any control questions is disqualified from participating in further tasks. In total, 9 pilot studies were released to recruit qualified annotators. For the final study, we use a different batch of control questions (i.e.\ there is no overlap in control questions between the pilot and final study). Workers with an accuracy of below 70\% were removed and prevented from doing more tasks.

\paragraph{Annotator Compensation}
Annotators were paid USD\$2.70 for each HIT. During the pilot study, workers were paid the base rate of \$1 and a bonus of \$1.70 if they passed the quality control. On average, a HIT took approximately 7 minutes to complete, resulting in a salary rate of \$15 per hour. In the final study, all workers were compensated for their time, irrespective of whether they passed the quality control.

\begin{figure}[t]
    \centering
    \includegraphics[width=\linewidth]{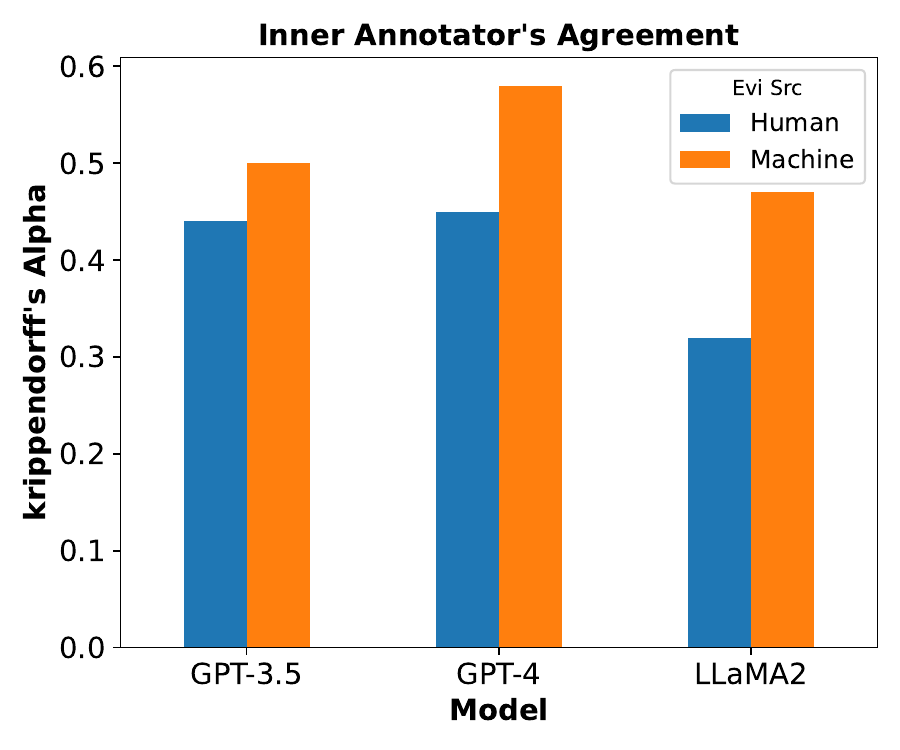}
    \caption{The inter-annotator agreement for the human annotations. ``Evi Src'' indicates the whether the evidence is \textit{Human}- or \textit{Machine}-selected.}
    \label{fig:human_iaa}
\end{figure}

\paragraph{Annotator Agreement}
Each claim was annotated by 5 annotators. As agreement metrics like kappa~\citep{McHugh-2012-kappa} are not applicable to our multi-label scenario, we use Krippendorff's alpha~\citep{krippendorff-2011-agreement} to compute annotator agreement.\footnote{For Krippendorff's alpha,
we use Jaccard distance to measure distance between two annotations (recall that each annotation result is a set, and as such we use Jaccard distance to measure the degree of set overlap).} \Cref{fig:human_iaa} presents the inter-annotator agreement results. On average, \four explanations produce the best agreement, followed by \threefive. \llamatwo has the lowest agreement, because its generated explanations tend to be quite noisy (we will revisit this in the next section). Interestingly, explanations generated using human-selected evidence have a lower agreement score. Overall, these agreement results indicate moderate to good agreement for \four and \threefive.

\subsection{Human Evaluation Results}\label{subsec:human_results}

Given the human annotation, we now compute citation recovery performance (set precision, recall, and F1 in \Cref{sec:transparency}) to assess attribution quality in the generated explanations. Results are presented in \Cref{tab:transparency_result}. \four has the best performance by a comfortable margin, suggesting that its generated explanations cite their sources more accurately than the other LLMs.
Manual analysis reveals that \llamatwo explanations often contain incorrect citations, such as \ex{[1][2][3]} which are directly copied from the prompt/instruction, as well as random links. That said, even the best model (\four with machine-selected evidence) only produces an F1 of approximately 0.74, which means a good proportion of citations are still not faithful. There's also substantial variance ($\pm$0.31), indicating the performance is far worse in the worst case scenario.
\begin{table}[!t]
    \centering
    \resizebox{\columnwidth}{!}{
   \begin{tabular}{cccccccc}
   \toprule
    \textbf{Evi Src} & \textbf{Gen} & \textbf{Precision} & \textbf{Recall} & \textbf{F1}\\
   \midrule
   \multirow{3}{3em}{\textit{Human}} 
   & GPT-4 & \textbf{0.62$\pm$0.29} & \textbf{0.67$\pm$0.29} & \textbf{0.63$\pm$0.29} \\
   & GPT-3.5 & 0.52$\pm$0.29 & 0.59$\pm$0.30 & 0.52$\pm$0.29 \\ 
   & LLaMA2 & 0.48$\pm$0.31 & 0.52$\pm$0.32 & 0.49$\pm$0.31\\
   \midrule
   \multirow{3}{3em}{\textit{Machine}} 
   & GPT-4 & \textbf{0.72$\pm$0.32} & \textbf{0.79$\pm$0.31} & \textbf{0.74$\pm$0.31} \\
   & GPT-3.5 & 0.55$\pm$0.39 & 0.55$\pm$0.39 & 0.53$\pm$0.37\\ 
   & LLaMA2 & 0.49$\pm$0.39 & 0.51$\pm$0.40 & 0.49$\pm$0.38\\
   \bottomrule
   \end{tabular}
   }
   \caption{Human evaluation results of generated explanations. The ``Evi Src'' column indicates whether evidence is \textit{Human}- or \textit{Machine}-selected.}
    \label{tab:transparency_result}
\end{table}

Looking at human-selected vs.\ machine-selected evidence for a given LLM, we generally see similar attribution performance, implying that we may not need perfectly curated evidence for generating explanations. To understand how different the machine-selected evidence is to the human-selected evidence, we compute set precision, recall and F1 using the human-selected evidence as ground truth, and present the results in Table \ref{tab:retrieval_f1}. Interestingly, across all LLMs we see only see a low to moderate F1 (0.3--0.5), meaning the machine-selected evidence is indeed quite different to human-selected evidence, but despite this difference, the resulting generated explanations do not differ much in terms of attribution quality.

\begin{table}[t]
    \centering
    \small
    \begin{tabular}{cccc}
    \toprule
    \textbf{Retriever} & \textbf{Precision} & \textbf{Recall} & \textbf{F1} \\
    \midrule
    GPT-4 & \textbf{0.40$\pm$0.21} & \textbf{0.75$\pm$0.23} & \textbf{0.47$\pm$0.20} \\
    GPT-3.5 & 0.34$\pm$0.22 & 0.60$\pm$0.35  & 0.39$\pm$0.22\\
    LLaMA2-70b & 0.29$\pm$0.18 & 0.68$\pm$0.32 & 0.36$\pm$0.18\\
    \bottomrule
    \end{tabular}
    \caption{Evidence retrieval performance.}
    \label{tab:retrieval_f1}
\end{table}

\section{Automatic Evaluation of Evidence Attribution}
We now attempt to automate the annotation process, i.e.\ automate the citation recovery task in the assessment protocol (\Cref{sec:transparency}). We experiment with two approaches: Natural Language Inference (NLI), and LLM annotators.

\paragraph{NLI}\label{para:pairwise_nli}
Most existing work frames the evaluation of citation quality as an NLI task~\citep{gao-etal-2023-enabling}. That is, given a pair of (premise, hypothesis), the NLI model outputs a label indicating whether the premise entails or contradicts the hypothesis. In our scenario, the premise is an evidence passage ($e_k$) while the hypothesis is an explanation sentence (a sentence in $X$). Given a selected evidence passage, we pair it with every sentence in the explanation and use a pretrained NLI model to detect instances of entailment: any sentence that is predicted to be entailed by the evidence passage is added to the recovered set ($X_{pred}^{mask}$).

In terms of pretrained NLI models, we use two models: \debertavthree~\citep{debertav3} and \true~\citep{honovich-etal-2022-true-evaluating}.\footnote{Detailed model versions: \debertavthree = deberta-v3-base, \true = t5\_xxl\_true\_nli\_mixture.}
The output label set for \debertavthree is ``entailment'', ``neutral'', and ``contradiction'', and for \true is ``1'' (entailment) and ``0'' (not entailment). For our experiments we are only concerned with the entailment class.

\paragraph{LLM} \label{para:llm_annotator}
We also experiment with using an LLM to do the citation recovery task. We zero-shot prompt an LLM to recover the sentences ($X_{pred}^{mask}$), using an instruction similar to what we provide to the human annotators (\Cref{sec:anno_detail}). For the input, we include the evidence passage ($e_k$) and all explanation sentences in the prompt. See Appendix \Cref{tab:prompt_exp_eval} for an example prompt.

In terms of LLMs, we use the same set of models that we used for generating explanations (i.e.\ \four, \threefive and \llamatwo), and further include several smaller LLMs: \llamathreeone~\citep{llama3.1}, \mistral~\citep{mistral} and \gemma~\citep{gemma}.\footnote{Detailed LLM versions: \llamathreeone = Llama-3.1-8B-Instruct, \mistral = Mistral-7B-Instruct-v0.3, \gemma = gemma-1.1-7b-it. All LLMs have at least 8K context window, and as such they fit the full prompt containing instructions, evidence set, and explanation.}

\paragraph{Evaluation Setting} We perform automatic annotation under two settings: (1) \textit{sampling}: for each claim, only one evidence passage is selected for masking -- this is the approach we used in the human annotation in \Cref{sec:human-eval}; and (2) \textit{full}: for each claim, we mask and evaluate all evidence passages, one at a time. The \textit{full} setting means we are assessing the attribution accuracy for every sentence that contains a citation marker in the explanation, and providing a more complete picture of the quality of the generated explanations (noting that this wasn't feasible with human evaluation). Note that
in this setting, for each claim we have multiple set precision, recall and F1 scores (one for each evidence passage; \Cref{para:attr_score}), and we aggregate them by computing the mean.

\begin{figure}[t]
    \centering
    \includegraphics[width=\linewidth]{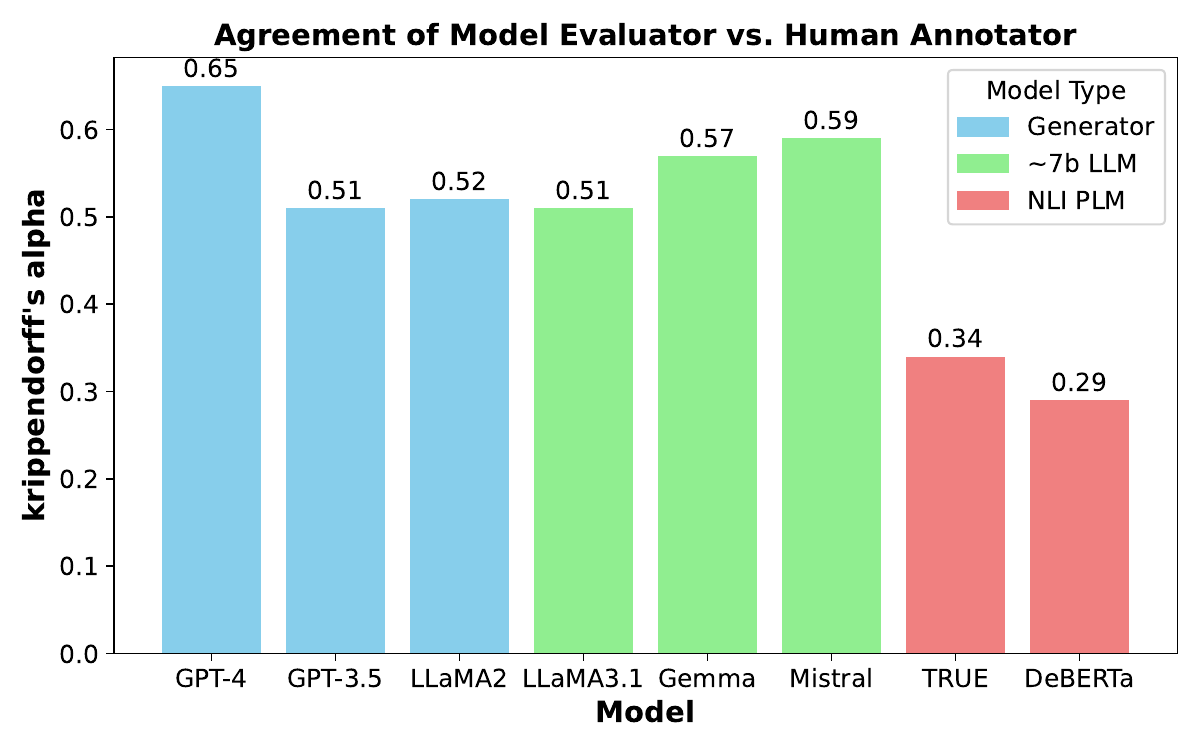}
    \caption{The agreement between automatic annotators vs.\ human annotators. ``Generator'' = models that are also used as (explanation) generators; ``$\sim$7b LLM'' refers to open-source LLMs with around 7B parameters; ``NLI PLM'' indicates PLMs used to conduct pairwise NLI, as described in \Cref{para:pairwise_nli}.}
    \label{fig:autoeval_iaa}
\end{figure}

\subsection{Agreement with Human Annotation}
\label{sec:agreement-with-human}

To understand how reliable LLMs and NLI models are as automatic annotators, we first assess their agreement with human annotations. 
We compute the agreement between automated and human annotators using krippendorff's alpha in a similar manner to that described in \Cref{subsec:human_results}. Here, however, we compute the agreement between two sets of annotation results: (1) an automatic annotator's results (e.g.\ \four's); and (2) aggregate results from all human annotators.\footnote{That is, we take the union of all recovered sentences ($X_{pred}^{mask}$) for each human annotator into a larger set.} We also standardise any labels outside the ground truth set ($X_{cit}^{mask}$) to a common label, as we are mostly interested in understanding the extent to which the automatic annotator selects the ground truth label when human annotators do so.\footnote{For example, if the ground truth set ($X_{cit}^{mask}$) is \{9\}, and the predicted set ($X_{pred}^{mask}$) from the automatic and human (grouped) annotator is \{7, 9\} and \{5, 9\} respectively, we will post-process both predicted sets into \{-2, 9\}.}

We present the agreement results in \Cref{fig:autoeval_iaa}. 
We can see LLM annotators have a much higher correlation with human annotators than NLI annotators. Due to the nature of the NLI input, the hypothesis is an explanation sentence that is taken out of context of the whole explanation, and may not always be interpretable. The LLM also assesses each explanation sentence individually, but they do this by taking into account the whole explanation context (as it is provided as part of the prompt). Given these results, we focus on using LLMs as annotators for the remainder of our experiments.

\subsection{Automatic Evaluation Results} 
We now present citation recovery results using LLMs as annotators for \textit{sample} in \Cref{tab:f1_vs_gt_sample_f1only} and \textit{all} in \Cref{tab:f1_vs_gt_all_f1only}. For the former we also include the human annotator results for comparison.

\paragraph{Comparing Automatic Annotator vs.\ Human Annotator} 
In \Cref{tab:f1_vs_gt_sample_f1only}, we can see that most automatic annotators have high agreement with human annotators: \four has the best attribution quality, followed by \threefive and \llamatwo. That said, if we look at the magnitude of the attribution quality, \threefive seems to overestimate this (i.e.\ its F1 numbers are much higher than those for humans), and this correlates with the agreement performance we saw in \Cref{sec:agreement-with-human} where \threefive has a lower agreement. \four produces results that best align with human annotators, and this is also reflected in \Cref{sec:agreement-with-human} where \four has the highest agreement among all models.

\begin{table}[!t]
    \resizebox{\columnwidth}{!}{
    \centering
   \begin{tabular}{ccccccccccc}
   \toprule
    \multirow{2}{*}{\diagbox[dir=NW,width=1.8cm]{\textbf{Ann}}{\textbf{Gen}}} & \multirow{2}{*}{\textbf{Evi Src}} & \textbf{GPT-4} & \textbf{GPT-3.5} & \textbf{LLaMA2} \\
     &  & \multicolumn{3}{c}{\textbf{Attribution F1}}\\
    \midrule
     \multirow{3}{*}{GPT-4} & \textit{Human} &  0.69 & 0.60 &0.50 \\
     & \textit{Machine} & 0.62 & 0.57 &  0.55 \\
     \midrule
     \multirow{3}{*}{GPT-3.5} & \textit{Human} &  0.83 & 0.73 & 0.61 \\
     & \textit{Machine} & 0.75 & 0.69 &  0.60 \\
     \midrule
     \multirow{3}{*}{LLaMA2} & \textit{Human} & 0.65 & 0.60 & 0.47 \\
     & \textit{Machine} & 0.57 & 0.51 &  0.47 \\
     \midrule
     \multirow{3}{*}{LLaMA3.1} & \textit{Human} &  0.57 & 0.49 & 0.45 \\
     & \textit{Machine} & 0.56 & 0.46 &  0.40 \\
     \midrule
     \multirow{3}{*}{Gemma} & \textit{Human} &  0.69 & 0.58 & 0.49 \\
     & \textit{Machine} & 0.59 & 0.52 &  0.46 \\
     \midrule
     \multirow{3}{*}{Mistral} & \textit{Human} &  0.58 & 0.49 & 0.42 \\
     & \textit{Machine} & 0.49 & 0.51 &  0.44 \\
     \midrule
     \multirow{3}{*}{Human} & \textit{Human} &  0.63 & 0.52 & 0.49 \\
     & \textit{Machine} & 0.74 & 0.53 &  0.49 \\
     \bottomrule
   \end{tabular}
   }
   \caption{Evidence attribution results for the \textit{sample} setting. The ``Evi Src'' column indicates whether evidence is \textit{Human}- or \textit{Machine}-selected, ``Ann'' the automatic annotators and ``Gen'' the explanation generators. }
    \label{tab:f1_vs_gt_sample_f1only}
\end{table}

\begin{table}[!t]
    \centering
    \resizebox{\columnwidth}{!}{
   \begin{tabular}{ccccccccccc}
   \toprule
    \multirow{2}{*}{\diagbox[dir=NW,width=1.8cm]{\textbf{Ann}}{\textbf{Gen}}} & \multirow{2}{*}{\textbf{Evi Src}} & \textbf{GPT-4} & \textbf{GPT-3.5} & \textbf{LLaMA2} \\
     &  & \multicolumn{3}{c}{\textbf{Attribution F1}}\\
     \midrule
     \multirow{3}{*}{GPT-4} & \textit{Human} &  0.67 & 0.61 & 0.50 \\
     & \textit{Machine} & 0.60 & 0.50 & 0.40 \\
     \midrule
     \multirow{3}{*}{GPT-3.5} & \textit{Human} & 0.80 & 0.75 &	0.64 \\
     & \textit{Machine} & 0.74 & 0.57 & 0.45 \\
     \midrule
     \multirow{3}{*}{LLaMA2} & \textit{Human} & 0.59 & 0.58 & 0.48 \\
     & \textit{Machine} & 0.53 & 0.41 &  0.34 \\
     \midrule
     \multirow{3}{*}{LLaMA3.1} & \textit{Human} &  0.47 & 0.47 & 0.40 \\
     & \textit{Machine} & 0.44 & 0.36 & 0.26 \\
    \midrule
     \multirow{3}{*}{Gemma} & \textit{Human} &  0.67 & 0.63 & 0.50 \\
     & \textit{Machine} & 0.55 & 0.47 &  0.34 \\
     \midrule
     \multirow{3}{*}{Mistral} & \textit{Human} &  0.54 & 0.49 & 0.39 \\
     & \textit{Machine} & 0.48 & 0.40 &  0.30 \\
     \bottomrule
   \end{tabular}
   }
   \caption{Evidence attribution results for the \textit{full} setting.}
    \label{tab:f1_vs_gt_all_f1only}
\end{table}

\paragraph{Comparing \textit{sample} vs.\ \textit{full}}

Looking at both \Cref{tab:f1_vs_gt_sample_f1only} and \Cref{tab:f1_vs_gt_all_f1only}, we see a similar trend across both \textit{sample} and \textit{full}, where \four is consistently the best generator, followed by \threefive and \llamatwo. Interestingly though, if we look at the impact of evidence source (``human'' vs.\ ``machine''), for \textit{sample} we see a small improvement from ``machine'' to ``human'', but this gap widens further in \textit{full} (for generator \threefive and \llamatwo in particular), suggesting that human-selected evidence does produce better explanations. This observation contrasts with what we saw previously in \Cref{subsec:human_results} (where machine-selected evidence is on-par with human-selected evidence), and shows that it is important to assess attribution for every sentence in the explanation to get the complete picture. 

To provide a more intuitive understanding of the overall evidence attribution quality of the generated explanations, we present \Cref{tab:transparency_proportion} to show the proportion of explanations where every sentence attributes the source accurately. We observe a large gap between ``human'' and ``machine'' across all generators, as well as all automatic annotators. This again reinforces that human-curated evidence leads to more transparent explanations.

\begin{table}[!t]
    \resizebox{\columnwidth}{!}{
    \centering
   \begin{tabular}{ccccccccccc}
   \toprule
    \multirow{2}{*}{\diagbox[dir=NW,width=1.8cm]{\textbf{Ann}}{\textbf{Gen}}} & \multirow{2}{*}{\textbf{Evi Src}} & \textbf{GPT-4} & \textbf{GPT-3.5} & \textbf{LLaMA2} \\
     &  & \multicolumn{3}{c}{\textbf{Fully Attributed Proportion}}\\
    \midrule
     \multirow{3}{*}{\four} & \textit{Human} &  31\% & 23\% & 17\% \\
     & \textit{Machine} & 9\% & 5\% & 4\% \\
     \midrule
     \multirow{3}{*}{\threefive} & \textit{Human} & 66\% & 50\% & 44\% \\
     & \textit{Machine} & 35\% & 28\% & 16\% \\
     \midrule
     \multirow{3}{*}{\llamatwo} & \textit{Human} & 45\% & 41\% & 31\% \\
     & \textit{Machine} & 24\% & 14\% &  6\% \\
     \midrule
     \multirow{3}{*}{\llamathreeone} & \textit{Human} &  34\% & 26\% & 25\% \\
     & \textit{Machine} & 13\% & 9\% &  5\% \\
     \midrule
     \multirow{3}{*}{\gemma} & \textit{Human} &  24\% & 14\% & 9\% \\
     & \textit{Machine} & 5\% & 2\% &  3\% \\
     \midrule
     \multirow{3}{*}{\mistral} & \textit{Human} &  19\% & 13\% & 11\% \\
     & \textit{Machine} & 8\% & 5\% &  1\% \\
     \bottomrule
   \end{tabular}
   }
   \caption{The proportion of Explanation where all sentences are ``Fully Attributed''. We use an attribution F1 threshold of 0.6 to classify an instance as ``Fully Attributed''. The ``Evi Src'' indicates whether evidence is \textit{Human}- or \textit{Machine}-selected, ``Ann'' the automatic annotators and ``Gen'' the explanation generators. }
    \label{tab:transparency_proportion}
\end{table}

\paragraph{Discussion}

Putting these results together, we can conclude as follows: (1) \four is the best model for generating fact-checking explanations based on evidences; (2) there is still much room for improvement for explanation generation, because even with human-curated evidence, only 31\% of \four's explanations attribute the sources accurately; (3) \four is the best automatic annotator for assessing evidence attribution; and (4) the selection of evidence is crucial, and we found that human-selected evidence tends to produce higher quality explanations across different LLM generators.

\section{Conclusion}
In this paper, we explore evaluating evidence attribution for fact-checking explanations. We implement a novel protocol using both human annotation and automated methods, and show that our LLM-based annotations have stronger correlation with human annotations compared to NLI models. While the best-performing LLM shows promising results, there is still substantial room for improvement in explanation generation. Our findings highlight the importance of evidence selection, as human-curated evidence tends to produce more attributable explanations. Ultimately, our research provides insights into enhancing AI trustworthiness and paves the way for future developments of transparent explanation generation for fact-checking.

\section*{Limitations}
This work only evaluates the evidence attribution of generated explanations, i.e.\ how faithful they are based on the sources. It is important to note that this evaluation does not assess factuality, which examines whether an explanation is factually correct. That said, there are already many studies on factuality, and our evaluation is orthogonal to these efforts. Additionally, while we tried our best to optimize the prompts, there remains the question of whether evidence attribution can be further improved with more prompt engineering. 
The PolitiHop dataset may contain out-of-date information, given it was developed a few years ago. As such, this may conflict with the LLM's parametric knowledge for certain claims. We saw this in our analyses where LLMs ignore evidence passages and generate statements based on their intrinsic knowledge. Also, the claims and evidence passages in PolitiHop are generally short single sentences. In practice, claims and evidence passages are likely to be longer.

\section*{Ethical Statement}
All experiments were conducted under the approval of an internal ethic committee (ethics application ID: 27463). We provide workers with a plain language statement and consent form detailing the research goal, methods, compensation, and potential risks (e.g.\ misleading information). Workers are free to stop any time once they started and they will still be compensated.

\section*{Acknowledgments}
We thank our reviewers for their valuable reviews
and feedback. We also thank our annotators for their contribution. This research was supported by The University of Melbourne’s Research Computing Services and the Petascale Campus Initiative.

\bibliography{anthology_revised,anthology_p2_revised,custom_revised}
\clearpage
\newpage
\appendix

\begin{table*}[!t]
    \scriptsize
    \centering
   \begin{tabular}{ccccccccccc}
   \toprule
    \multirow{2}{*}{\diagbox[dir=NW,width=1.3cm]{\textbf{Eval}}{\textbf{Gen}}} & \multirow{2}{*}{\textbf{Evi Src}} & \multicolumn{3}{c}{\textbf{GPT-4}} & \multicolumn{3}{c}{\textbf{GPT-3.5}} & \multicolumn{3}{c}{\textbf{LLaMA2-70B}} \\
     &  & \textbf{Precision} & \textbf{Recall} & \textbf{F1} & \textbf{Precision} & \textbf{Recall} & \textbf{F1} & \textbf{Precision} & \textbf{Recall} & \textbf{F1} \\
    \midrule
    \multirow{3}{*}{GPT-4} & \textit{Human} & 0.59±0.32 & 0.96±0.18 & 0.69±0.25 & 0.50±0.32 &  0.88±0.30 & 0.60±0.29 & 0.40±0.33 & 0.76±0.43 &	0.50±0.34 \\
     & \textit{Machine} & 0.50±0.30 & 0.97±0.17 & 0.62±0.26 & 0.47±0.32 & 0.87±0.33 & 0.57±0.30 & 0.47±0.37 &  0.79±0.39 & 0.55±0.35 \\
     \midrule
     \multirow{3}{*}{GPT-3.5} & \textit{Human} & 0.82±0.34 & 0.88±0.32 & 0.83±0.32 & 0.72±0.39 &  0.79±0.38 & 0.73±0.37 & 0.59±0.45 & 0.68±0.47 &	0.61±0.45 \\
     & \textit{Machine} & 0.72±0.39 & 0.83±0.38 & 0.75±0.37 & 0.71±0.42 & 0.72±0.42 & 0.69±0.4 & 0.59±0.44 &  0.67±0.46 & 0.60±0.43 \\
     \midrule
     \multirow{3}{*}{LLaMA2} & \textit{Human} & 0.59±0.4 & 0.76±0.42 & 0.65±0.39 & 0.54±0.37& 0.74±0.42 & 0.6±0.36 & 0.4±0.37	&0.61±0.49&0.47±0.39 \\
     & \textit{Machine} & 0.49±0.35&0.75±0.43&0.57±0.35 & 0.48±0.41&0.62±0.47&0.51±0.4 & 0.41±0.37&0.61±0.48&0.47±0.39 \\
     \midrule
     \multirow{3}{*}{LLaMA3.1} & \textit{Human} & 0.47±0.31 & 0.80±0.40 & 0.57±0.32 & 0.41±0.33 & 0.70±0.45 & 0.49±0.34 & 0.36±0.31 & 0.66±0.47 & 0.45±0.35 \\
     & \textit{Machine} & 0.46±0.32 & 0.83±0.37 & 0.56±0.31 & 0.38±0.33 & 0.66±0.46 & 0.46±0.34 & 0.31±0.28 & 0.65±0.47 & 0.40±0.33 \\
     \midrule
     \multirow{3}{*}{Gemma} & \textit{Human} & 0.61±0.35 & 0.86±0.34 & 0.69±0.33 & 0.51±0.37 & 0.75±0.42 & 0.58±0.36 & 0.42±0.38 & 0.67±0.47 & 0.49±0.39 \\
     & \textit{Machine} & 0.52±0.38 & 0.78±0.41 & 0.59±0.37 & 0.47±0.39 & 0.68±0.45 & 0.52±0.38 & 0.42±0.41 & 0.59±0.48 & 0.46±0.40 \\
     \midrule
     \multirow{3}{*}{Mistral} & \textit{Human} & 0.46±0.27 & 0.90±0.29 & 0.58±0.26 & 0.38±0.28 & 0.82±0.37 & 0.49±0.28 & 0.32±0.25 & 0.76±0.42 & 0.42±0.28 \\
     & \textit{Machine} & 0.37±0.26 & 0.86±0.35 & 0.49±0.27 & 0.39±0.26 & 0.85±0.34 & 0.51±0.25 & 0.34±0.28 & 0.77±0.41 & 0.44±0.29 \\
     \midrule
     \multirow{3}{*}{Human} & \textit{Human} & 0.62±0.29  & 0.67±0.29  & 0.63±0.29 & 0.52±0.29 & 0.59±0.30 & 0.52±0.29 & 0.48±0.31 & 0.52±0.32 & 0.49±0.31 \\
     & \textit{Machine} & 0.72±0.32 & 0.79±0.31 & 0.74±0.31 & 0.55±0.39 & 0.55±0.39 & 0.53±0.37 & 0.49±0.39 & 0.51±0.40 & 0.49±0.38 \\
     \bottomrule
   \end{tabular}
   \caption{Full Evidence attribution results for the \textit{sample} setting. The ``Evi Src'' column indicates whether evidence is \textit{Human}- or \textit{Machine}-selected, ``Ann'' the automatic annotators and ``Gen'' the explanation generators. }
    \label{tab:f1_vs_gt_sample}
\end{table*}

\begin{table}[!t]
    \small
    \centering
    \resizebox{\columnwidth}{!}{
   \begin{tabular}{cccccccc}
   \toprule
    \textbf{Evi Src} & \textbf{Gen}  & \textbf{Utility} & \textbf{Entropy}\\
   \midrule
   \multirow{3}{3em}{\textit{Human}} 
   & \four  & \textbf{0.28$\pm$0.18}  & 66.86$\pm$19.38 \\
   & \threefive & 0.41$\pm$0.18 & \textbf{72.47$\pm$19.52} \\ 
   & \llamatwo & 0.34$\pm$0.16 & 65.13$\pm$18.00\\
   \midrule
   \multirow{3}{3em}{\textit{Machine}} 
   & \four & \textbf{0.18$\pm$0.19} &  \textbf{76.34$\pm$17.88} \\
   & \threefive & 0.21$\pm$0.18 & 70.47$\pm$20.89\\ 
   & \llamatwo  & 0.21$\pm$0.17 & 66.20$\pm$21.84\\
   \bottomrule
   \end{tabular}
   }
   \caption{Human evaluation results of Utility and Entropy in generated explanations. The ``Evi Src'' column indicates whether evidence is \textit{Human}- or \textit{Machine}-selected.}
    \label{tab:transparency_utility}
\end{table}

\begin{table}[t]
    \centering
    \small
    \resizebox{\columnwidth}{!}{
    \begin{tabular}{cccc}
    \toprule
    \textbf{Evi Src} & \textbf{Gen} & $\mathbf{F1=1.0}$ & $\mathbf{F1\geq0.9}$ \\
    \midrule
    \multirow{3}{3em}{\textit{Human}}
    & \four & \textbf{83.43$\pm$3.66} & \textbf{74.90$\pm$15.62} \\
    & \threefive & 61.11$\pm$30.57 & 73.22$\pm$25.75 \\
    & \llamatwo & 63.65$\pm$25.14 & 73.60$\pm$17.89 \\
    \midrule
    \multirow{3}{3em}{\textit{Machine}}
    & \four & \textbf{84.09$\pm$7.94} & \textbf{83.88$\pm$7.86} \\
    & \threefive & 83.62$\pm$6.74 & 83.15$\pm$6.75 \\
    & \llamatwo & 82.96$\pm$6.90 & 82.96$\pm$6.90 \\
    \bottomrule
    \end{tabular}
    }
    \caption{Utility score after filtering via $F1$ threshold, 
    scores are presented in the form of ave$\pm$std. The ``Evi Src'' column indicates whether evidence is \textit{Human}- or \textit{Machine}-selected.}
    \label{tab:filered_utility}
\end{table}

\begin{table*}[!t]
    \scriptsize
    \centering
   \begin{tabular}{ccccccccccc}
   \toprule
    \multirow{2}{*}{\diagbox[dir=NW,width=1.3cm]{\textbf{Eval}}{\textbf{Gen}}} & \multirow{2}{*}{\textbf{Evi Src}} & \multicolumn{3}{c}{\textbf{GPT-4}} & \multicolumn{3}{c}{\textbf{GPT-3.5}} & \multicolumn{3}{c}{\textbf{LLaMA2-70B}} \\
     &  & \textbf{Precision} & \textbf{Recall} & \textbf{F1} & \textbf{Precision} & \textbf{Recall} & \textbf{F1} & \textbf{Precision} & \textbf{Recall} & \textbf{F1} \\
    \midrule
    \multirow{3}{*}{GPT-4} & \textit{\textit{Human}} & 0.56±0.31 & 0.96±0.19 & 0.67±0.25 & 0.52±0.33 & 0.88±0.31 & 0.61±0.30 & 0.41±0.33 & 0.77±0.41 &	0.50±0.33 \\
     & \textit{Machine} & 0.50±0.32 & 0.95±0.21 & 0.60±0.28 & 0.41±0.33 & 0.76±0.41 & 0.50±0.32 & 0.33±0.35 & 0.61±0.48 & 0.40±0.36 \\
     \midrule
     \multirow{3}{*}{GPT-3.5} & \textit{\textit{Human}} & 0.78±0.37 & 0.84±0.36 & 0.80±0.35 & 0.75±0.39 &  0.78±0.39 & 0.75±0.38 & 0.63±0.45 & 0.68±0.46 &	0.64±0.44 \\
     & \textit{Machine} & 0.70±0.39 & 0.82±0.38 & 0.74±0.37 & 0.58±0.46 & 0.61±0.46 & 0.57±0.44 & 0.44±0.46 & 0.50±0.49 & 0.45±0.46 \\
     \midrule
     \multirow{3}{*}{LLaMA2} & \textit{\textit{Human}} & 0.53±0.39&0.71±0.45&0.59±0.4 & 0.53±0.39&0.68±0.45&0.58±0.39 & 0.43±0.4&0.59±0.49&0.48±0.41 \\
     & \textit{Machine} & 0.47±0.38&	0.67±0.47&0.53±0.4 & 0.38±0.4&	0.5±0.48&0.41±0.41 & 0.30±0.37&	0.43±0.49&0.34±0.40 \\
    \midrule
     \multirow{3}{*}{LLaMA3.1} & \textit{\textit{Human}} & 0.44±0.44 & 0.55±0.49 & 0.47±0.44 & 0.46±0.45 & 0.53±0.48 & 0.47±0.44 & 0.38±0.44 & 0.46±0.49 & 0.40±0.44 \\
     & \textit{Machine} & 0.40±0.43 & 0.52±0.50 & 0.44±0.44 & 0.35±0.44 & 0.40±0.47 & 0.36±0.43 & 0.25±0.40 & 0.29±0.45 & 0.26±0.40 \\
     \midrule
     \multirow{3}{*}{Gemma} & \textit{\textit{Human}} & 0.60±0.37 & 0.83±0.37 & 0.67±0.35 & 0.57±0.38 & 0.79±0.39 & 0.63±0.36 & 0.45±0.40 & 0.65±0.47 & 0.50±0.40 \\
     & \textit{Machine} & 0.48±0.39 & 0.74±0.44 & 0.55±0.39 & 0.44±0.41 & 0.60±0.47 & 0.47±0.40 & 0.31±0.41 & 0.44±0.49 & 0.34±0.41 \\
     \midrule
     \multirow{3}{*}{Mistral} & \textit{\textit{Human}} & 0.41±0.26 & 0.89±0.31 & 0.54±0.26 & 0.37±0.26 & 0.81±0.37 & 0.49±0.28 & 0.29±0.26 & 0.69±0.45 & 0.39±0.29 \\
     & \textit{Machine} & 0.37±0.27 & 0.85±0.36 & 0.48±0.28 & 0.31±0.29 & 0.69±0.45 & 0.40±0.31 & 0.22±0.26 & 0.57±0.49 & 0.30±0.30 \\
     \bottomrule
   \end{tabular}
   \caption{Full Evidence attribution results for the \textit{full} setting. The ``Evi Src'' column indicates whether evidence is \textit{\textit{\textit{\textit{Human}}}}- or \textit{\textit{Machine}}-selected, ``Ann'' the automatic annotators and ``Gen'' the explanation generators. }
    \label{tab:f1_vs_gt_all}
\end{table*}

\begin{table*}[!t]
    \small
    \centering
   \begin{tabular}{p{2 \columnwidth}}
   \toprule
    \textbf{Prompt} \\
   \midrule 
    Instructions: You are required to write an accurate, coherent and logically consistent explanation for the claim based on the given veracity and list of reasons in one paragraph. Use an unbiased and journalistic tone. When citing several search results, use [1][2][3]. Ensure that each reason is cited only once. Do not cite multiple reasons in a single sentence. \newline
    \newline
    Reasons:\newline
    Reason [1] What’s more, the picture referenced in the Facebook post alleging that anglerfish are typically 7 feet is taken from the Australian Museum’s 2012 exhibit titled  "Deep Oceans.
    Reason [2] When the exhibit opened in June 2012, The Sydney Morning Herald reported on how the exhibit’s team had created an "oversized anglerfish" and listed the many steps in making it: "Pieces such as the oversized anglerfish, with huge fangs and antenna-like flashing rod to attract prey, begin with cutting and welding a metal frame, then sculpting material over it and, finally, hand painting it," the story says.\newline
    \newline
    Claim:The typical anglerfish is seven feet long.\newline
    Veracity: False\newline
    Explanation:\\
    \bottomrule
   \end{tabular}
   \caption{Prompts for explanation generation}
    \label{tab:prompt_exp_gen}
\end{table*}




\begin{table*}[!t]
    \small
    \centering
   \begin{tabular}{p{2 \columnwidth}}
   \toprule
    \textbf{Prompt} \\
   \midrule 
    Find the most suitable explanation sentence(s) that can cite the given reason sentence. Return the sentence number(s) separated by comma, e.g., 0 or 0,2. Return -1 if no suitable sentences are found. Consider semantic citation relationships, not just keyword matching. Only return numbers, DO NOT include any additional output.\newline
    \newline
    Reason Sentence:\newline
    What’s more, the picture referenced in the Facebook post alleging that anglerfish are typically 7 feet is taken from the Australian Museum’s 2012 exhibit titled  "Deep Oceans.\newline
    \newline
    Explanation Sentences:\newline
    1. The claim that the typical anglerfish is seven feet long is false.\newline
    2. One reason for this is that the picture used in the Facebook post to support the claim is taken from the 2012 exhibit titled \"Deep Oceans\" at the Australian Museum, which featured an \"oversized anglerfish\" [8].\newline
    3. This suggests that the anglerfish showcased in the exhibit is not typical in size.",\newline
    4. Furthermore, an article from The Sydney Morning Herald that reported on the exhibit mentioned the many steps involved in creating the exhibit's \"oversized anglerfish\" [10]."\newline
    5. These findings indicate that the anglerfish portrayed in the picture and the exhibit are not representative of the typical size of anglerfish.\newline
    \newline
    Answers: \\
    \bottomrule
   \end{tabular}
   \caption{Prompts for automatic explanation evaluation}
    \label{tab:prompt_exp_eval}
\end{table*}

\section{Assessing Perceived Utility}
Although not the main focus of this paper, in addition to Evidence Attribution, we also assess Explanation Utility as a complementary metric. Utility captures whether users find an explanation helpful in clarifying the claim~\cite{liu-etal-2023-evaluating, gao-etal-2023-enabling}. 
\paragraph{Utility Evaluation}\label{sub:utility} Utility measures to what extent the generated explanation clarifies a claim. Though a five-point Likert scale is commonly used~\citep{liu-etal-2023-evaluating,gao-etal-2023-enabling}, \citet{ethayarajh-jurafsky-2022-authenticity} found that averaging using the Likert scale can result in a biased estimate. As such we use Direct Assessment~\citep{graham-etal-2013-continuous} where annotators rate on a (continuous) sliding scale from $0$--$100$, where $100$ = best. We ask annotators the following question: \textit{How helpful is the explanation in clarifying the truthfulness of the claim?}. We aggregate the judgements of multiple annotators for each explanation by computing the mean. We also applied a Bayesian model for utility score calibration, but it showed a similar tendency, so we decided to use the original scores. 

\paragraph{Utility Score Calibration}\label{sec:util_calib}
Perceiving the utility of fact-checking explanation is a subjective task. On the one hand, annotators might disagree on how useful the explanation is. Some workers may consistently provide low utility scores for all explanations due to their high standards, while others might be more lenient. Additionally, certain workers may only utilize a narrow range of the scoring scale such as the central part. On the other hand, while deploying our task on AMT provides a convenient and cost-effective solution, it comes with challenges such as high variance between workers, poor calibration, and the potential to draw misleading scientific conclusions~\citep{karpinska-etal-2021-perils}. 

Motivated by the aforementioned reasons, we used a simple Bayesian model~\citep{mathur-etal-2018-towards} to calibrate the annotated utility scores. The calibration functions as follows: assuming the utility score $s$ is normally distributed around the true utility $\mu$ of the explanation, we use an accuracy parameter $\tau$ to model each worker's accuracy: higher value indicates smaller errors. The full generative modelling works as follows:
\begin{itemize}
\item For each explanation $i$, we draw true utility $\mu_i$ from the standard normal distribution.
\begin{equation}\label{eqn:exp_true_utility}
    \mu_i \sim \mathcal{N}(0,1)
\end{equation}

\item Then for each annotator $j$, we draw their accuracy $\tau$ from a shared Gamma prior with shape parameter $k$ and rate parameter $\theta$ \footnote{We use $k=1.5$ and $\theta=0.5$ based on our manual inspection of preliminary experiments.}.
\begin{equation}\label{eqn:worker_acc}
    \tau_j \sim \mathcal{G}(k,\theta)
\end{equation}

\item The annotator’s utility score $s_{i,j}$ is then drawn from a normal distribution with mean $\mu_i$ and accuracy $\tau_j$.
\begin{equation}\label{eqn:final_utility}
    s_{i,j} \sim \mathcal{N}(\mu_i,\tau^{-1})
\end{equation}
\end{itemize}

Our goal is to maximize the likelihood of the observation of annotated utility score:
\begin{equation}\label{eqn:likelihood}
\begin{aligned}
     P(s) = \int_{j} P(\tau_j) \int_i P(\mu_i)P(s_{i,j}|\mu_i,\tau)d\tau d\mu &
    \\
     = \int_j \Gamma (\tau_j|k,\theta) \int_i \mathcal{N} (\mu_i|0,1)& \\ \mathcal{N}(s_{i,j}|\mu_j,\tau_j^{-1}) d\tau d\mu&
\end{aligned}
\end{equation}

We first standardize individual annotators’ utility scores via z-scoring to enhance comparability and reduce potential biases. Afterwards we use Expectation Propagation~\citep{minka-2001-expectation} to infer posterior over true utility score $\mu$ and annotator accuracy $\tau$~\footnote{We implemented the model with \href{https://dotnet.github.io/infer/}{Infer.NET} framework~\citep{InferNET18}.}.

\paragraph{Utility Results} High utility doesn't necessarily imply high evidence attribution.
Although we found a a general pattern between utility and evidence attribution (Table \ref{tab:transparency_utility}), \threefive achieved the highest utility score when using human-selected evidence (72.47), even though its attribution score is much lower compared to \four (0.52 vs.\ 0.63). After filtering for transparent explanation with a threshold of $F1=1.0$ and $F1\geq0.9$, \four achieved the highest utility score, as shown in Table~\ref{tab:filered_utility}.This indicates that high utility \threefive explanations are not attributable. Overall, this demonstrates that utility and citation represent two distinct qualities and has an important implication: an explanation that appears helpful can actually still be misleading.

\section{Annotation Entropy}\label{sec:prf1_formula}
We used another metric Entropy~\citep{shannon-entropy} to measure the randomness and the degree of uncertainty in the system:
\begin{equation}\label{eqn:entropy}
    H = -\sum {{p_k}\;{log}\;{p_k}}
\end{equation}
In multi-label scenario, the entropy of the label probability distribution reflects the likelihood of each chosen label. It also influences the probability of agreement on the label~\citep{marchal-etal-2022-establishing}. For instance, consider a certain annotation result $[0, 0, 4, 0]$, which represents the occurrences of each option, annotators exhibit high agreement in choosing the $2^{nd}$ sentence to cite the reason (index starts from 0). In contrast, $[1, 1, 1, 1]$ shows evenly distributed choices on each option, which suggests greater uncertainty among annotators. Consequently, the latter will have higher entropy. In our task, we utilize entropy as an indicator of annotation uncertainty. We compute the normalized probability of each claim annotation and then apply equation~\ref{eqn:entropy} to calculate entropy.

\section{Prompts}
We developed and optimized prompt for our task. We present evidence selection prompt in~\Cref{tab:prompt_evi_sel}, explanation generation prompt in~\Cref{tab:prompt_exp_gen} and explanation evaluation prompt for LLM as annotators in~\Cref{tab:prompt_exp_eval}.

\section{Experiment Details}
We ran all offline models on 4 Nvidia A100 GPUs in a data parallel fashion. Explanation generation with \four and \threefive takes around 2 hours and \llamatwo takes around 4 hours. The cost of generating explanations was \$49 (US Dollar) with \four and \$10 (US Dollar) with \threefive. We present full evidence attribution results for \textit{sample} setting in \Cref{tab:f1_vs_gt_sample} and \textit{full} setting in \Cref{tab:f1_vs_gt_all}.

\begin{table*}[!t]
    \small
    \centering
   \begin{tabular}{p{2 \columnwidth}}
   \toprule
    \textbf{Prompt} \\
   \midrule 
    Instructions: You are required to retrieve a subset of reasons from the provided full reasons. The sentences in this subset should be coherent and logically consistent, presenting the most crucial information necessary to establish the veracity of the claim. Aim for the minimum number of sentences in the subset while maintaining the completeness and clarity. When extract reasons, use [1,2,3]. At last, provide a justification explaining why they are good reasons and how they form a logically consistent reasoning process.
    \newline
    Demonstration:\newline
    Reasons:\newline
    Reason [0]: Anglerfish may have a reputation for being among the creepier-looking ocean-dwellers, but it’s not because they grow to be seven feet long, as a viral image on Facebook claims.\newline
    Reason [1]: The Jan. 12 post shows a young girl reaching toward what appears to be a very large anglerfish mounted on display at a museum.\newline
    Reason [2]: The text above the image reads, "So,... I’ve spent my entire life thinking the Deep Sea Angler Fish was about the size of a Nerf football.\newline
    Reason [3]: What’s more, the picture referenced in the Facebook post alleging that anglerfish are typically 7 feet is taken from the Australian Museum’s 2012 exhibit titled  "Deep Oceans".\newline
    Reason [4]: The anglerfish in the photo is actually a large-scale sculpture model of the fish made of plaster.\newline
    Reason [5]: When the exhibit opened in June 2012, The Sydney Morning Herald reported on how the exhibit’s team had created an "oversized anglerfish" and listed the many steps in making it: "Pieces such as the oversized anglerfish, with huge fangs and antenna-like flashing rod to attract prey, begin with cutting and welding a metal frame, then sculpting material over it and, finally, hand painting it," the story says. \newline
    \newline
    Claim: The typical anglerfish is seven feet long.
    Veracity: False\newline
    Extracted Reasons: [3,5]\newline
    Justification: Reason [3] establishes that the Facebook post's claim relies on a picture from the Australian Museum's 2012 exhibit. Reason [5] then reveals that the anglerfish in the exhibit is an oversized sculpture, not an actual specimen. Together, these reasons logically demonstrate that the viral claim of typical anglerfish being seven feet long is false, as it is based on a misrepresented image from an exhibit.\newline
    \newline
    Here's the actual task:\newline
    Reasons:\newline
    Reason [0]: Amid fears about the coronavirus disease, a YouTube video offers a novel way to inoculate yourself: convert to Islam.\newline
    Reason [1]: "20m Chinese gets converted to Islam after it is proven that corona virus did not affect the Muslims," reads the title of a video posted online Feb. 18.\newline
    Reason [2]: The footage shows a room full of men raising an index finger and reciting what sounds like the Shahadah, a statement of faith in Islam.\newline
    Reason [3]: That’s because the footage is from at least as far back as May 26, 2019, when it was posted on Facebook with this caption: "Alhamdulillah welcome to our brothers in faith."\newline
    Reason [4]: On Nov. 7, 2019, it was posted on YouTube with this title: "MashaaAllah hundreds converted to Islam in Philippines."\newline
    Reason [5]: Both posts appeared online before the current outbreak of the new coronavirus, COVID-19, was first reported in Wuhan, China, on Dec. 31, 2019.\newline
    Reason [6]: But even if the footage followed the outbreak, Muslims are not immune to COVID-19, as the Facebook post claims.\newline
    Reason [7]: After China, Iran has emerged as the second focal point for the spread of COVID-19, the New York Times reported on Feb. 24.\newline
    Reason [8]: "The Middle East is in many ways the perfect place to spawn a pandemic, experts say, with the constant circulation of both Muslim pilgrims and itinerant workers who might carry the virus."\newline
    Reason [9]: On Feb. 18, Newsweek reported that coronavirus "poses a serious risk to millions of inmates in China’s Muslim prison camps."\newline
    Claim: It was stated on February 18, 2020 in a YouTube post: “20 million Chinese converted to Islam after it’s proven that the coronavirus doesn’t affect Muslims.”\newline
    Veracity: False\newline
    Extracted Reasons:\\
    \bottomrule
   \end{tabular}
   \caption{Prompts for evidence selection.}
    \label{tab:prompt_evi_sel}
\end{table*}

\section{Scientific Artifacts}
We list the licenses of different artifacts used in this paper: PolitiHop (MIT),\footnote{\url{https://github.com/copenlu/politihop}} Huggingface Transformers (Apache License 2.0).\footnote{\url{https://github.com/huggingface/transformers}} Our source code and annotated data are released under MIT license.

\clearpage

\clearpage
\newpage
\section{Annotation Interface} \label{app:interface}

\noindent\begin{minipage}[H]{\textwidth}
    \centering
    \includegraphics[page=1,scale=1, width=1\textwidth]{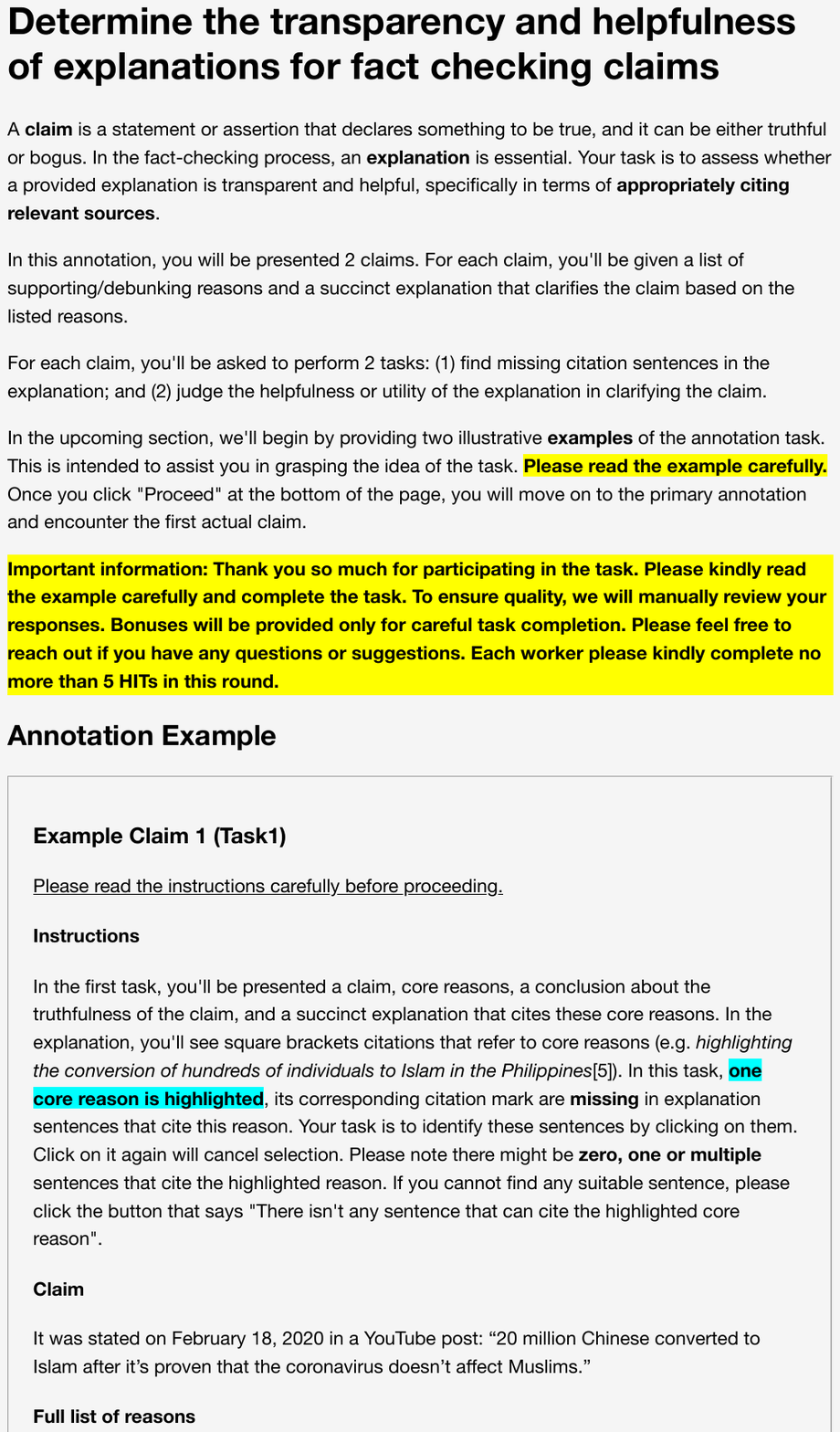}
    \captionof{figure}{Annotation Interface Page 1}
\end{minipage}

\clearpage
\newpage
\noindent\begin{minipage}[H]{\textwidth}
    \centering
    \includegraphics[page=1,scale=1, width=1\textwidth]{figs/annotation_interface_p.pdf}
    \captionof{figure}{Annotation Interface Page 2}
\end{minipage}

\clearpage
\newpage
\noindent\begin{minipage}[H]{\textwidth}
    \centering
    \includegraphics[page=2,scale=1, width=1\textwidth]{figs/annotation_interface_p.pdf}
    \captionof{figure}{Annotation Interface Page 3}
\end{minipage}

\clearpage
\newpage
\noindent\begin{minipage}[H]{\textwidth}
    \centering
    \includegraphics[page=3,scale=1, width=1\textwidth]{figs/annotation_interface_p.pdf}
    \captionof{figure}{Annotation Interface Page 4}
\end{minipage}

\clearpage
\newpage
\noindent\begin{minipage}[H]{\textwidth}
    \centering
    \includegraphics[page=4,scale=1, width=1\textwidth]{figs/annotation_interface_p.pdf}
    \captionof{figure}{Annotation Interface Page 5}
\end{minipage}

\clearpage
\newpage
\noindent\begin{minipage}[H]{\textwidth}
    \centering
    \includegraphics[page=5,scale=1, width=1\textwidth]{figs/annotation_interface_p.pdf}
    \captionof{figure}{Annotation Interface Page 6}
\end{minipage}

\clearpage
\newpage
\noindent\begin{minipage}[H]{\textwidth}
    \centering
    \includegraphics[page=6,scale=1, width=1\textwidth]{figs/annotation_interface_p.pdf}
    \captionof{figure}{Annotation Interface Page 7}
\end{minipage}

\end{document}